\documentclass[11pt]{article}

\usepackage[margin=1in]{geometry}
\usepackage{amsmath,amssymb}
\usepackage{booktabs}
\usepackage{graphicx}
\usepackage{tikz}
\usetikzlibrary{arrows.meta,calc,positioning}
\usepackage{hyperref}
\usepackage{cleveref}
\usepackage{xcolor}
\definecolor{accent}{RGB}{0,120,130}
\title{Do Active SAE Feature Planes Carry More Holonomy?\\ A Preregistered Reversal in Gemma}
\author{Larry Richards\\Independent Researcher}
\date{\today}

\newcommand{\glossentry}[2]{%
  \par\medskip\noindent\phantomsection\label{gloss:#1}\textbf{#2.}\ %
}

\begin{document}

\maketitle

\begin{abstract}
This paper tests whether holonomy concentrates on active sparse-autoencoder (SAE) feature planes in Gemma 2 2B, a concrete operationalization of the broader semantic-concentration prediction. Holonomy is measured at the final-token layer-12 to layer-13 residual-stream readout by carrying a local frame around small loops using the instrument's restricted-Jacobian transport rule, then normalizing the resulting rotation by enclosed area. The design, materiality threshold, analysis, and verdict rules were preregistered and frozen before the analysed measurements were inspected. The prediction was falsified in reverse: active-feature planes carried less holonomy than matched mixed-feature controls, with an adjusted log contrast of \(-0.29439\) and 95\% interval \([-0.43989,-0.14889]\). A magnitude-only explanation was not supported in this design, while the three-way ordering across random, mixed-feature, and active-feature planes was undefined at matched magnitude because common support failed. Post-freeze diagnostics at the same readout supported the area law on a small validation subset, bounded matched-center displacement under a simple paired regression, and identified transport distortion as a live mechanism or confound. The result is therefore a narrow, auditable operational reversal, not a causal claim that meaning suppresses holonomy. The cause remains open, with activation-strength geometry, degree of feature engagement, dictionary geometry, matched-center displacement, activation-manifold proximity, and transport shear as live alternatives.
\end{abstract}

\section{Introduction}
% JOB: motivate the object and stake the refuted hypothesis honestly.
This paper tests a concrete \emph{semantic-concentration prediction}: that holonomy concentrates on active SAE feature planes, whose directions are selected from active sparse-autoencoder features at the base point.

Were the semantic-concentration prediction to hold, holonomy would distinguish semantically dense regions from sparse ones --- a geometric signature of where meaning concentrates. This is the motivating possibility, stated conditionally.

Measuring holonomy of a learned representation is not itself new: it was introduced for vision networks by Sevetlidis and Pavlidis \cite{sevetlidis2026gauge}, who also noted that their loop planes need not align with meaning. In their setting, features are raw layer activations or PCA directions; here, features are learned SAE directions. This paper's contribution is the first, to our knowledge, holonomy measurement inside a transformer language model; it takes up the semantic question they left open, using an instrument and planes built from SAE features, under a preregistered confirmatory test.

Every analytic choice --- the hypothesis, the materiality threshold, the design, the analysis --- was fixed before the data were inspected.

As a consequence, this operationalized semantic-concentration prediction was reversed, not merely unmet: active-feature planes carried materially less holonomy than the matched mixed-feature controls.

Holonomy is measurable by the instrument defined in \Cref{sec:geometric-setup-instrument}. Active-feature planes carried less holonomy, not more; the underlying mechanism remains open.

\section{Concepts and the Problem}
% ORDER RATIONALE: concepts first (the objects), then the motivated problem (why the design assembles them this way). Logical motivation throughout, never chronology (D4): tell why the design must be this way, not how we arrived at it.
% SCOPE BOUNDARY: Section 2 MOTIVATES (why three plane types, why these hypotheses, holonomy intuition); Section 5 SPECIFIES (exact plane construction, formal hypotheses, the 3x2x390 structure); Section 4 is the holonomy INSTRUMENT (the math/analytic measurement, and invariances invoked in situ). Section 2 must not duplicate Section 4/5 content.
% SCOPE BOUNDARY: invariances do not belong in Section 2 beyond the optional 2.2 intuition; they are a Section 4 topic, invoked at the operative step.

\subsection{The setting: activation space and SAE features}
% JOB: introduce the substrate and the feature objects. Activation space / residual stream (the space the model computes in). Then the motivation chain for SAE features: raw directions are polysemantic -> an SAE disentangles them into roughly-monosemantic learned directions -> an "SAE feature" is one such direction. Hands off the term "SAE feature" to all later chapters. Concept register; promote comprehension-critical terms from the glossary (D7). Use D13 vocabulary.
A transformer processes text into high-dimensional vectors that pass through its layers \cite{vaswani2017attention}. At a fixed layer, these vectors live in a shared activation space, or residual stream; each point is one activation. This is the space whose geometry this paper studies.

\begin{figure}[ht]
\centering
\begin{tikzpicture}[
  font=\small,
  box/.style={draw=black!55, rounded corners=2pt, line width=0.45pt, fill=black!2},
  softbox/.style={box, fill=black!1},
  highlight/.style={draw=accent, fill=teal!10, rounded corners=2pt, line width=0.8pt},
  arrow/.style={-{Latex[length=2.2mm,width=1.6mm]}, line width=0.55pt, draw=black!65},
  accentarrow/.style={-{Latex[length=2.4mm,width=1.8mm]}, line width=0.8pt, draw=accent},
  label/.style={font=\footnotesize, text=black!75},
  note/.style={font=\footnotesize, text=black!70, align=center}
]

% Column anchors
\coordinate (L) at (0,0);
\coordinate (M) at (4.15,0);
\coordinate (R) at (8.55,0);

% Column headers
\node[font=\bfseries\small, align=center] at ($(L)+(0,2.85)$) {SAE coordinate system};
\node[font=\bfseries\small, align=center] at ($(M)+(0,2.85)$) {decoder matrix};
\node[font=\bfseries\small, align=center] at ($(R)+(0,2.85)$) {activation space};

% Left: feature list
\node[box, minimum width=2.25cm, minimum height=4.75cm, anchor=center] (featurebox) at (L) {};
\node[label, anchor=west] (f1) at ($(featurebox.north west)+(0.25,-0.45)$) {feature 1};
\node[label, anchor=west] (f2) at ($(featurebox.north west)+(0.25,-0.95)$) {feature 2};
\node[label, anchor=west] (dotsA) at ($(featurebox.north west)+(0.25,-1.50)$) {$\vdots$};
\node[highlight, minimum width=1.62cm, minimum height=0.42cm, anchor=west] (fiBox) at ($(featurebox.north west)+(0.20,-2.15)$) {};
\node[font=\footnotesize, text=accent, anchor=west] (fi) at ($(featurebox.north west)+(0.30,-2.15)$) {feature $i$};
\node[label, anchor=west] (dotsB) at ($(featurebox.north west)+(0.25,-2.78)$) {$\vdots$};
\node[label, anchor=west] (f16k) at ($(featurebox.south west)+(0.25,0.48)$) {feature 16k};
\node[note] at ($(featurebox.south)+(0,-0.50)$) {feature named by\\SAE coefficient};

% Middle: decoder matrix
\node[box, minimum width=1.57cm, minimum height=4.15cm, anchor=center] (matrixbox) at (M) {};
\foreach \y in {1.62,1.18,0.74,0.30,-0.58,-1.02,-1.46} {
  \draw[black!25, line width=0.35pt] ($(matrixbox.west)+(0,\y)$) -- ($(matrixbox.east)+(0,\y)$);
}
\node[highlight, minimum width=1.45cm, minimum height=0.48cm, anchor=center] (rowi) at ($(matrixbox.center)+(0,-0.12)$) {};
\node[font=\footnotesize, text=accent, align=center] at (rowi.center) {row $i$};

% Right: activation-space sphere and direction
\coordinate (sphereCenter) at (R);
\def\spherer{0.92cm}
\draw[black!45, line width=0.45pt] (sphereCenter) circle (\spherer);
\draw[black!25, line width=0.35pt, dashed] ($(sphereCenter)+(-\spherer,0)$) arc[start angle=180,end angle=360,x radius=\spherer,y radius=0.25cm];
\draw[black!45, line width=0.40pt] ($(sphereCenter)+(\spherer,0)$) arc[start angle=0,end angle=180,x radius=\spherer,y radius=0.25cm];
\draw[black!25, line width=0.35pt, dashed] ($(sphereCenter)+(0,-\spherer)$) arc[start angle=-90,end angle=90,x radius=0.28cm,y radius=\spherer];
\draw[black!45, line width=0.40pt] ($(sphereCenter)+(0,\spherer)$) arc[start angle=90,end angle=270,x radius=0.28cm,y radius=\spherer];
\node[label, anchor=south] at ($(sphereCenter)+(0,1.22)$) {2304 dimensions};
\fill[black!65] (sphereCenter) circle (1.35pt);
\draw[accentarrow] (sphereCenter) -- ++(0.63,0.67) coordinate (dirEnd);
\node[font=\footnotesize, text=accent, anchor=west, align=left] at ($(dirEnd)+(0.08,0.00)$) {$d_i$};
\node[note] at ($(sphereCenter)+(0,-1.70)$) {feature $i$'s direction\\in activation space};

% Bridges
\draw[accentarrow] ($(fiBox.east)+(0.05,0)$) -- node[above, label, yshift=0.04cm] {select} ($(rowi.west)+(-0.05,0)$);
\draw[accentarrow] ($(rowi.east)+(0.05,0)$) -- node[above, label, yshift=0.16cm, pos=0.42] {decode} ($(sphereCenter)+(-1.38,0.02)$);
\node[font=\scriptsize, text=black!70, align=center] at (6.25,-0.58)
  {$\textcolor{accent}{d_i} = \sum_{k=1}^{2304} W_{i,k}\,e_k$};

\end{tikzpicture}
\caption{\textbf{The decoder bridge.} Each SAE feature corresponds to one row of the decoder matrix --- its direction in activation space. A feature can be named by its coefficient (left) or its decoder direction (right).}
\label{fig:decoder-bridge}
\end{figure}
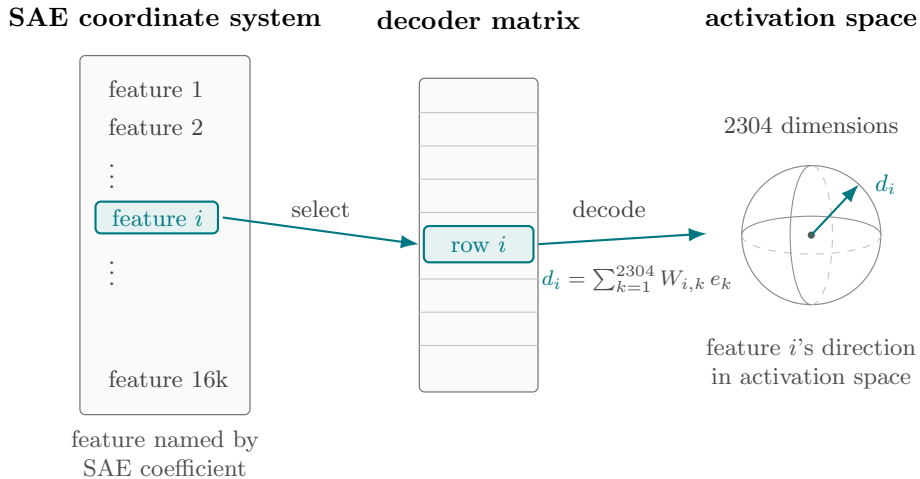

Individual directions in activation space are hard to interpret, because a single direction tends to superpose several unrelated meanings. This property is called \emph{polysemanticity} \cite{elhage2022toy}. A sparse autoencoder (SAE) is a learned tool that addresses this: it learns its own set of approximately monosemantic features, each identified by a coefficient \cite{bricken2023towards,lieberum2024gemma}. These features live in the SAE's own coordinate system, but the SAE's decoder matrix bridges the two: each feature is one row of that matrix, which is the feature's equivalent direction in activation space (\Cref{fig:decoder-bridge}). An SAE feature can therefore be named in either system --- by its coefficient in the SAE, or by its decoder direction in activation space.

For a given residual-stream activation, the SAE assigns each feature a coefficient, and only a few are positive: those features are active at that point, and the rest are inactive. Which features are active varies from one residual-stream activation to the next.

These features are the ``meaningful directions'' the rest of the paper refers to; when the paper builds planes, it uses their decoder directions in activation space.

\subsection{Holonomy, conceptually}
% JOB: the intuition of holonomy in words only (the operational/mathematical instrument is deferred to Section 4). A frame carried around a small closed loop returns rotated; that rotation is holonomy; it is a signature of curvature. Establish that holonomy is a property of a plane through a point (which sets up why plane types matter). Optional single intuitive sentence: the twist depends on the loop itself, not on how it is traced -- stated as intuition only, no invariance machinery here (invariances are invoked in situ in Section 4, at the step each one serves). Hands off to Section 4.
Holonomy is a geometric quantity measured at a point in activation space. Loosely, it captures how much the space twists in a small region: a larger value means more geometric structure.

We measure it within a two-dimensional plane through the point, and its value depends on which plane is chosen. A plane's orientation is fixed by a choice of two directions in activation space --- for instance, two SAE feature directions, or two picked at random. The choice of directions, and why three kinds of plane are needed, are taken up in Section~2.4.

The precise definition, and the instrument that measures it, are given in \hyperref[sec:geometric-setup-instrument]{Section 4}.

\subsection{The question}
% JOB: state the motivated problem. If meaning has geometric structure, holonomy might concentrate on meaningful directions -- the semantic-concentration prediction (introduced properly here; it was named in the Introduction). State the stake briefly (what it would give: distinguishing semantically dense regions from sparse ones), and why it is worth testing. Conditional mood for the benefit (the prediction failed).
Activation space has meaningful directions --- the SAE features of Section 2.1 --- and holonomy measures geometric structure on planes through a point. Is holonomy larger on (a) planes built from semantically selected directions, or (b) planes built without regard to meaning?

This is the \emph{semantic-concentration prediction}: that holonomy concentrates on meaningful directions. Were it to hold, holonomy could distinguish semantically dense regions from sparse ones, helping chart the terrain of activation space.

The contrast is in how a plane's directions are \emph{chosen}, not what they happen to contain. To test the prediction, then, we compare planes built under different selection rules, as the next section explains.

\subsection{Three plane types}
% JOB: motivate the three-way split as logical necessity, not by decree. Reasoning chain: a feature plane alone says nothing without a comparison -> a control that removes meaning while matching size isolates meaning (the mixed-feature plane) -> but a control could differ for geometric reasons unrelated to meaning -> a random floor calibrates the geometric baseline. Three roles: signal (active-feature), control (mixed-feature), floor (random). Note: this is also where the confound logic is first felt ("a control could differ for geometric reasons") -- the seed of the Section 7 caveat. D13 names; D4 (no chronology).
To test the prediction, we compare holonomy across planes whose directions are chosen by different rules. We use three.

\begin{itemize}
\item The \textbf{active-feature plane} is the signal: its directions are selected for semantic content, so if holonomy concentrates on meaningful directions, it should be largest here.
\item The \textbf{mixed-feature plane} is the near comparison: still feature-based and tied to the same base point, but no longer the full signal plane.
\item The \textbf{random plane} is the far comparison: its directions are chosen with no reference to SAE features, giving a geometric floor where nothing is selected for meaning.
\end{itemize}

We make two comparisons, at two distances, because a single comparison could not tell whether a difference reflects semantic selection or some incidental geometric property of the comparison plane. The exact construction of the three plane types is given in the Experimental Design; whether even this design fully separates meaning from incidental geometry is taken up in the \hyperref[sec:discussion]{Discussion}.

\subsection{Three competing hypotheses}
% JOB: motivate the hypotheses as answers to rival explanations. If holonomy tracks meaning, feature planes carry more (H-sem); but maybe it merely tracks size, so that rival must be tested (H-mag); and if structure truly orders, it should rise from random to control to feature (H-grad). Three hypotheses because three rival explanations must be separated. Hands off to Section 5 (formal hypothesis statements) and Section 3 (the verdict machinery that adjudicates them). Plane types (2.4) precede hypotheses (2.5) because the hypotheses are stated in terms of the plane types.
The prediction that holonomy concentrates on meaningful directions is one explanation for a difference between plane types, but not the only one. To test it honestly, we state it alongside the rival explanations it must be separated from. Three hypotheses result.

The first is \textbf{H-sem}, the semantic main effect and the study's primary prediction, which the design was powered to test. It is the semantic-concentration prediction stated as a testable hypothesis: holonomy tracks meaning, so active-feature planes carry more of it than the controls. This is the prediction the study sets out to test, now in a form a verdict rule can decide.

The second is \textbf{H-mag}, a rival that has nothing to do with meaning. In the Experimental Design we define in-plane magnitude, a planar property whose value varies candidate by candidate, independent of holonomy. \textbf{H-mag}, the magnitude main effect, is the hypothesis that the differences in holonomy size are explained by in-plane magnitude alone. Because this rival must be ruled out, the plane types are compared at a common in-plane magnitude; the details are given in the Experimental Design.

The third is \textbf{H-grad}, which concerns the order of the three types. If holonomy genuinely reflects degree of meaningful structure, it should not merely separate active-feature planes from the rest but rise across all three: random lowest, mixed-feature between, active-feature highest. \textbf{H-grad}, the monotone gradient, is the hypothesis that holonomy increases monotonically from random to mixed to active.

Three hypotheses, then, because three distinct explanations must be told apart: that holonomy tracks meaning (\textbf{H-sem}), that it tracks magnitude (\textbf{H-mag}), and that it orders the plane types (\textbf{H-grad}). These are separate questions, each decided on its own, so one experiment yields a verdict on all three. The formal statement of each, and the rules that decide them, are given in the \hyperref[sec:methodology]{Methodology} and \hyperref[sec:experimental-design]{Experimental Design}.

\section{Methodology}
\label{sec:methodology}
% JOB: explain what was measured, what was controlled, and how the frozen analysis was run.
This study's central methodological commitment is preregistration: the hypotheses, the decision rule, the materiality threshold, the analysis, and the design were all fixed in writing and frozen before the analysed measurements were inspected. This chapter states the commitments; the specific design that instantiates them is given in the Experimental Design chapter.

This discipline is what makes the result trustworthy rather than merely surprising. Because the result reverses the semantic-concentration prediction the study set out to support, the natural suspicion is that the analysis was steered. Freezing every analytic choice before the data were visible removes the freedom that steering would require, so the result was determined by the design rather than chosen after the fact.

The analysis is single-pass: the data are analysed once, with no peeking at results before the analysis is fixed, and no result-dependent stopping or extension. There is no opportunity to keep looking until the answer changes.

Every prediction is decided by a verdict rule fixed in advance, by where its confidence interval falls relative to the materiality threshold, not by judgement applied after seeing the numbers. Each prediction receives one of four verdicts:

\begin{table}[ht]
\centering
\begin{tabular}{p{0.22\linewidth}p{0.68\linewidth}}
\hline
\textbf{Verdict} & \textbf{Condition} \\
\hline
Corroborated & confidence interval lies entirely above the materiality threshold \\
Falsified & confidence interval lies entirely below the materiality threshold \\
Inconclusive & confidence interval straddles the materiality threshold \\
Undefined & the required comparison is unsupported (no matched magnitude; see \Cref{sec:experimental-design}) \\
\hline
\end{tabular}
\end{table}

A verdict is never forced: where the comparison cannot be made, the prediction is left undefined rather than resolved.

Two kinds of selection happen before any holonomy is measured: which text passages from the corpus to use, and which candidate planes pass a pre-measurement geometric check. Both were applied before any holonomy was observed and calibrated without reference to outcomes. No measurement was excluded on the basis of its holonomy value; candidate planes were rejected only by the pre-measurement non-degeneracy floor.

The analysed data were verified to be byte-for-byte identical to the frozen measurement before any estimate was computed. The trustworthiness of the result therefore rests on a checkable fact, that the exact frozen data were analysed, not on an assurance.

Reproducibility has a specific, bounded basis. Model execution and the Jacobian-vector products run on Apple Silicon, the MPS backend, which is the sole backend of this study; the small pullback-metric matrices are then evaluated deterministically in CPU float64. Within that fixed backend, results are bitwise reproducible. No CUDA alternative was used, and no bitwise guarantee is claimed across other backends. The complete environment is committed in the repository's environment files: \path{.python-version}, \path{ENVIRONMENT.md}, \path{pyproject.toml}, and \path{uv.lock}.

The design was revised exactly once, before any experiment data were observed, to correct a specification error; the revision touched no experiment-sample base point, consumed only reserved pilot points, and was itself recorded and re-frozen. The specific correction is described in the \hyperref[sec:experimental-design]{Experimental Design} chapter.

The entire record, including the preregistration, its one revision, the analysis specification, and the measurement data, is committed with full version history and publicly auditable in the released experiment repository. Every freeze-before-data claim in this paper is checkable against that record rather than taken on the author's word.

\section{Geometric Setup and Instrument}
\label{sec:geometric-setup-instrument}
% JOB: define holonomy operationally and show the instrument is trustworthy.
The instrument measures holonomy by carrying a frame around a small closed loop in a two-dimensional affine plane and recording how the frame rotates when the loop closes. Two unit directions $d_1,d_2$ fix the plane's orientation. Collect them as columns of $D = \begin{bmatrix} d_1 & d_2 \end{bmatrix}$; their column space $C(D)$ is the through-origin plane of that orientation. To place that orientation somewhere in activation space, use an offset point $b$, giving the affine plane $b+C(D)$. In this experiment's measured loops, the chosen offset is the matched center $c$, constructed in \Cref{sec:experimental-design}. With that offset, a loop in the affine plane $c+C(D)$ has the form
\begin{equation}
\label{eq:loop}
  \gamma(t) = c + \rho\bigl(\cos 2\pi t\, d_1 + \sin 2\pi t\, d_2\bigr), \qquad t \in [0,1]
\end{equation}
The radius is calibrated from the raw base activation, $\rho = 6.0\times10^{-3}\,\lVert h\rVert$, keeping the probe small relative to the activation scale.

All geometry is the geometry induced by the model's downstream map, not ordinary Euclidean geometry. In this experiment, the downstream map \(F\) sends a patched final-token residual-stream activation at \path{model.model.layers[12]} to the final-token residual-stream output of \path{model.model.layers[13]}, with the original 64-token corpus prefix and attention mask held fixed. Let $J$ be the Jacobian of that map, evaluated through Jacobian--vector products. The measured quantity is therefore readout-dependent: it is a property of the layer-12 activation site together with this layer-12-to-layer-13 readout and the Euclidean inner product on the readout space, not an intrinsic property of activation space alone. At a point on the loop, the two plane directions push forward to the columns of the restricted Jacobian:
\begin{equation}
\label{eq:restricted-jacobian}
  JD = \begin{bmatrix} J d_1 & J d_2 \end{bmatrix}
\end{equation}
Define the plane's induced metric, the $2\times2$ pullback Gram matrix $M$, by
\begin{equation}
\label{eq:pullback-gram}
  M = JD^\top JD
\end{equation}
This matrix is what makes lengths, areas, and angles in the plane refer to the model's own geometry.

Transport carries the frame around the loop by composing these restricted Jacobians segment by segment. Writing $JD_i$ for the restricted Jacobian at the $i$-th loop point, each segment contributes $S_i = JD_{i+1}^{+} JD_i$, where ${}^{+}$ is the Moore--Penrose pseudoinverse, and the transport operator accumulates from the identity:
\begin{equation}
\label{eq:transport}
  T = \bigl(JD_n^{+} JD_{n-1}\bigr)\bigl(JD_{n-1}^{+} JD_{n-2}\bigr)\cdots\bigl(JD_1^{+} JD_0\bigr)
\end{equation}
When the loop closes, $T$ is the residual map the frame has undergone. This restricted-Jacobian, or extrinsic pullback, transport uses the Jacobians directly and never differentiates the metric or constructs Christoffel symbols. The pullback metric $M$ enters only through the enclosed area and the non-degeneracy floor below, not through transport. The accumulated operator is not assumed to be exactly orthogonal; shear and numerical non-orthogonality are allowed, and the reported angle is the antisymmetric rotational component extracted from the residual operator.

The rotation the frame picks up is the antisymmetric part of $T$. For a small residual map, this antisymmetric component is the first-order rotational generator; the symmetric component records stretch and shear rather than rotation. With $\Delta = T - I$:
\begin{equation}
\label{eq:theta}
  \theta = \tfrac{1}{2}\bigl(\Delta_{21} - \Delta_{12}\bigr)
\end{equation}
This is the effective rotation within the plane. The enclosed area is taken under the same pullback metric, using the restricted Jacobian at the loop's starting point:
\begin{equation}
\label{eq:area}
  A = \rho^2 \sqrt{\det M}
\end{equation}
This convention omits the constant disk factor \(\pi\).\footnote{Including \(\pi\) would multiply every area by the same constant, shifting all log holonomy responses by the same amount and leaving contrasts and verdicts unchanged.}
The response is holonomy magnitude, rotation per unit enclosed area:
\begin{equation}
\label{eq:response}
  H = \frac{\lvert\theta\rvert}{A}
\end{equation}
The orientation-independent magnitude $H$ is the primary response; the signed value $\theta/A$ is retained only as a diagnostic. (Numerical safeguards used in computing these quantities --- a pseudoinverse cutoff, a determinant floor inside the square root, and a small-area guard in the division --- are recorded with the source; they protect against degeneracy and do not change the quantities defined here.)

Dividing by enclosed area is load-bearing, not cosmetic. It makes the response a local curvature rather than a quantity that simply grows with loop size. Because the response is curvature per unit area, the comparison between plane types compares curvatures, not enclosed areas, which closes an obvious objection before it can be raised.

The instrument was not accepted on faith. Synthetic and oracle checks verified loop closure, area normalization, extraction of the rotational angle, and reversal of the signed twist when loop direction is reversed. A post-freeze validation at the same layer-12-to-layer-13 readout as the confirmatory experiment swept radii \(0.5\rho,\rho,2\rho\) over the first four frozen base points, active-feature and mixed-feature planes, at low matched magnitude. The pooled log--log slope of \(|\theta|\) versus \(\rho\) was \(2.03\), with the pooled fit combining loops with different intercepts; the eight per-loop slopes ranged from \(1.989\) to \(2.005\), supporting the area-law interpretation for this instrument at the measured site. This validation is reported as post-freeze instrument validation; it does not enter or modify the preregistered verdict.

Near-degenerate planes, where the two directions are almost parallel under the pullback metric, were excluded before any holonomy was measured. A single fixed rule, $\det M > 0.413$, was applied identically to all plane types, with the threshold calibrated blind on throwaway synthetic points. No plane was excluded on the basis of its holonomy value.

\section{Experimental Design}
\label{sec:experimental-design}
% JOB: show the test could have killed the hypothesis, and was frozen before data.
The study is a preregistered, single-pass factorial experiment. The full design, the decision rules, the materiality threshold, and the analysis were fixed and frozen in writing before the analysed measurements were inspected; nothing was tuned after the answer was visible.

The structure crosses three plane types with two magnitude levels, measured at each of $390$ base points, for $3 \times 2 \times 390 = 2{,}340$ measurements. Each base point is measured under all six combinations, so every comparison is made within a base point rather than across different ones.

\paragraph{System under study.} The model is Gemma~2 2B (\path{google/gemma-2-2b}) \cite{gemma2024gemma2}. All activations are taken at the residual stream after block~12, \path{resid_post}, the output of \path{model.model.layers[12]}, a $2304$-dimensional space. SAE feature directions come from the Gemma Scope sparse autoencoder for this site (\path{google/gemma-scope-2b-pt-res}, \path{layer_12/width_16k/average_l0_82}) \cite{lieberum2024gemma}, whose suffix denotes an average L0 of about $82$ active SAE features; the directions used to build planes are its normalized decoder rows. Base points are residual-stream activations drawn from WikiText-103 \cite{merity2017pointer} (the \path{wikitext-103-raw-v1} split at a pinned dataset revision) under a fixed seed: each passage is truncated to its first $64$ tokens, and the base point is the activation at the final token.

\paragraph{Constructing the plane types.} An \emph{active-feature plane} is spanned by two SAE features whose coefficients are positive at the base point, selected as the highest joint-activation-strength pair that remains non-degenerate under the pullback metric. A \emph{mixed-feature plane}, the control, pairs one active SAE feature with one inactive one (an SAE feature with zero code at that base point), the inactive partner drawn at random and the pair subject to the same non-degeneracy floor; it is constructed independently and does not reuse a direction from the active-feature plane. A \emph{random plane} is spanned by two random unit directions in activation space, not SAE feature directions, and serves as a geometric floor.

\paragraph{The size-confound problem.} Comparing holonomy across plane types requires first ruling out an innocent alternative: the plane types may differ not only in how their directions are chosen, but also in a kind of size. If holonomy simply tracks that size, then a difference between plane types would be a fact about size, not about the active-feature construction.

The remedy is the one used to compare apples and oranges when the apples sampled happen to be large and the oranges small, and sweetness varies with size. Comparing the piles directly would tell us little: any difference might be size, not fruit. Instead we compare apples and oranges \emph{of the same size}. Here the corresponding task is to compare planes of different types at the same size, so that a difference in holonomy cannot be blamed on size.

\paragraph{The canonical plane.} The loop in \Cref{eq:loop} uses the affine plane $c+C(D)$: the orientation $C(D)$ plus the matched center $c$. Magnitude matching uses the same orientation, but with the canonical offset $O$. Through the origin $O$, the column space itself is the canonical plane $\Pi_0 = C(D)$: every combination $\alpha d_1 + \beta d_2$, written $D\mathbf{x}$ with in-plane coordinates $\mathbf{x} = (\alpha,\beta)$.

The magnitude construction uses this canonical placement, $\Pi_0$. Meanwhile, the holonomy loop is drawn in $\Pi_c = c + C(D)$, the same column space translated to the loop's center $c$. Thus the same orientation is placed two ways: through $O$ to measure magnitude, and through the loop center to measure holonomy.

The activation $h$ does not, in general, lie in $\Pi_0$; it floats off the sheet. Dropping it onto $\Pi_0$ gives a shadow point $P$. The segment from $O$ to $P$ is the in-plane part, and the remaining perpendicular part is the out-of-plane part. The size used for matching is the length of that in-plane part: how far the activation's shadow reaches into the plane.

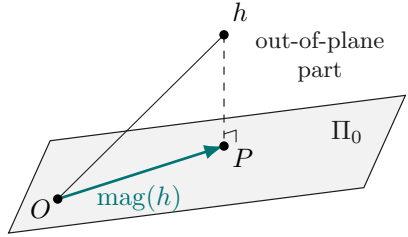
\begin{figure}[ht]
\centering
\begin{tikzpicture}[
  >=Latex,
  font=\small,
  sheet/.style={draw=black!90, fill=black!5, line width=0.45pt},
  guide/.style={draw=black!90, dashed, line width=0.45pt},
  accent/.style={draw=teal!90!black, line width=1pt},
  accent text/.style={teal!90!black},
  point/.style={circle, fill=black, inner sep=1.25pt}
]

  \coordinate (O) at (0.85,0.55);
  \coordinate (P) at (3.05,1.25);
  \coordinate (h) at (3.05,2.72);

  \coordinate (A) at (0.20,0.10);
  \coordinate (B) at (4.90,0.70);
  \coordinate (C) at (5.45,1.92);
  \coordinate (D) at (0.75,1.32);
  \filldraw[sheet] (A) -- (B) -- (C) -- (D) -- cycle;
  \node[black!90] at (4.68,1.44) {$\Pi_0$};

  \draw[black!90, line width=0.4pt] (O) -- (h);
  \draw[guide] (h) -- (P);
  \draw[accent, -{Latex[length=2.6mm]}] (O) -- (P);
  \node[accent text] at (1.93,0.56) {$\operatorname{mag}(h)$};

  \draw[black!90, line width=0.4pt] ($(P)+(0.16,0.05)$) -- ++(0,0.16) -- ++(-0.16,-0.05);

  \node[point, label={[below left=-2pt]$O$}] at (O) {};
  \node[point, label={[below right=-1pt]$P$}] at (P) {};
  \node[point, label={[above right=-1pt]$h$}] at (h) {};
  \node[black!90, align=center, font=\footnotesize] at ($(3.92,2.03)+(4mm,4mm)$) {out-of-plane\\part};
\end{tikzpicture}
\caption{\textbf{In-plane magnitude.} The activation $h$ is projected onto the canonical plane $\Pi_0=C(D)$ through the origin. The shadow segment $OP$ is the in-plane part, while the perpendicular segment is the out-of-plane part.}
\label{fig:in-plane-magnitude}
\end{figure}

\paragraph{In-plane magnitude.} This size has a name: the in-plane magnitude, written $\operatorname{mag}(h)$. It is the reach of the activation's shadow into the canonical plane, the length of the segment $OP$. When $\operatorname{mag}(h)$ is large, the activation leans strongly into the plane; when it is small, the activation barely reaches in. The geometry is shown in \Cref{fig:in-plane-magnitude}.

This value is measured per plane; it differs from one plane to another, and the design's magnitude levels are shared targets, denoted \(m\), derived from those per-plane measurements. To make the shadow length exact, we still need the model-induced ruler, introduced below.

\paragraph{Magnitude normalization.} The plane types differ in their typical $\operatorname{mag}(h)$: active-feature planes tend to be larger, random planes smaller, with mixed-feature planes in between. Measuring holonomy on the planes as they stand would confound this difference in $\operatorname{mag}(h)$, resulting in an unfair comparison.

To make the comparison fair, the design normalizes $\operatorname{mag}(h)$ before measuring holonomy. It fixes shared target magnitudes and rescales every plane to a target: a plane whose native $\operatorname{mag}(h)$ exceeds the target is brought down, and one whose native $\operatorname{mag}(h)$ falls short is brought up. Only after this normalization is the loop drawn and holonomy measured.

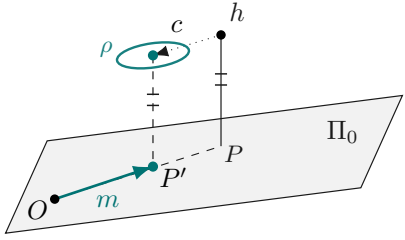
\begin{figure}[ht]
\centering
\begin{tikzpicture}[
  >=Latex,
  font=\small,
  sheet/.style={draw=black!90, fill=black!5, line width=0.45pt},
  guide/.style={draw=black!90, dashed, line width=0.45pt},
  ghost/.style={draw=black!90, dashed, line width=0.4pt},
  accent/.style={draw=teal!90!black, line width=1pt},
  accent text/.style={teal!90!black},
  point/.style={circle, fill=black, inner sep=1.25pt},
  accent point/.style={circle, fill=teal!90!black, inner sep=1.35pt}
]

\definecolor{Accent}{RGB}{24,128,128}

  \coordinate (O2) at (0.85,0.55);
  \coordinate (Pold) at (3.05,1.25);
  \coordinate (hghost) at (3.05,2.72);
  \coordinate (Pp) at (2.15,0.98);
  \coordinate (c) at (2.15,2.45);
  \coordinate (A2) at (0.20,0.10);
  \coordinate (B2) at (4.90,0.70);
  \coordinate (C2) at (5.45,1.92);
  \coordinate (D2) at (0.75,1.32);
  \filldraw[sheet] (A2) -- (B2) -- (C2) -- (D2) -- cycle;
  \node[black!90] at (4.66,1.44) {$\Pi_0$};

  % Old in-plane leg, ghosted, and new target leg.
  \draw[ghost] (O2) -- (Pold);
  \node[black!90, font=\footnotesize] at ($(Pold)+(0.17,-0.10)$) {$P$};
  \draw[accent, -{Latex[length=2.6mm]}] (O2) -- (Pp)
    node[pos=0.54, below=1pt, accent text] {$m$};

  % Preserved out-of-plane height.
  \draw[black!90, line width=0.45pt] (hghost) -- (Pold);
  \draw[guide] (c) -- (Pp);
  \draw[black!90, line width=0.55pt] ($(hghost)!0.35!(Pold)+(-0.08,0)$) -- ($(hghost)!0.35!(Pold)+(0.08,0)$);
  \draw[black!90, line width=0.55pt] ($(c)!0.35!(Pp)+(-0.08,0)$) -- ($(c)!0.35!(Pp)+(0.08,0)$);
  \draw[black!90, line width=0.55pt] ($(hghost)!0.45!(Pold)+(-0.08,0)$) -- ($(hghost)!0.45!(Pold)+(0.08,0)$);
  \draw[black!90, line width=0.55pt] ($(c)!0.45!(Pp)+(-0.08,0)$) -- ($(c)!0.45!(Pp)+(0.08,0)$);

  % Unlabeled ghost position of h and the in-plane center shift.
  \draw[black!90, dotted, -{Latex[length=2mm]}] (hghost) -- (c);

  % Loop in the translated plane through c.
  \draw[Accent, line width=0.9pt] (c) ellipse [x radius=0.48, y radius=0.16, rotate=8];
  \node[Accent, font=\footnotesize] at ($(c)+(-0.62,0.08)$) {$\rho$};

  \node[point, label={[below left=-2pt]$O$}] at (O2) {};
  \node[point, label={[above right=-1pt, black!90]$h$}] at (hghost) {};
  \node[accent point, label={[below right=-2pt]$P'$}] at (Pp) {};
  \node[accent point, label={[above right=4pt]$c$}] at (c) {};

\end{tikzpicture}
\caption{\textbf{Magnitude normalization.} To match a target magnitude $m$, the shadow endpoint is rescaled from $P$ to $P'$ along the same in-plane direction. The matched center $c$ is formed by lifting $P'$ by the unchanged out-of-plane part. The holonomy loop of radius $\rho$ is then drawn around $c$, using the same plane directions $d_1,d_2$.}
\label{fig:magnitude-normalization}
\end{figure}

Normalization is performed on the canonical plane $\Pi_0$. The activation's in-plane component, $\overrightarrow{OP}$, is rescaled along its own direction to a scalar target length $m$, while the out-of-plane component is preserved. The loop center is then $c$, the activation $h$ carried to where its shadow has length $m$: the rescaled in-plane part, lifted by the untouched out-of-plane part. The loop is drawn in $C(D)$ at this center (\Cref{fig:magnitude-normalization}). Its plane $\Pi_c = c + C(D)$ is a translation of the canonical plane $C(D)$: the same column space and radius, centered at $c$ rather than at $O$. Only the center and the in-plane length differ.

\paragraph{Matched magnitude and the fallback.} Common support is the overlap band in native $\operatorname{mag}(h)$ where the plane types being compared all have observations. The design first checks whether that band is wide enough to support shared target magnitudes. If it is, the targets are fixed before holonomy is measured by taking the 25th and 75th percentiles of the pooled native magnitudes within the overlap band.

In the frozen run, the three plane types did not share enough common magnitude range for a three-way match. Random planes occupied much lower native magnitudes than the feature-based planes, so forcing a three-way comparison would have been extrapolation rather than matching. The verdict-bearing comparison, active-feature versus mixed-feature, did have common support, and its shared targets were $m_{\mathrm{low}} = 32.28$ and $m_{\mathrm{high}} = 45.00$.

The preregistered fallback therefore applies: the active-feature versus mixed-feature comparison proceeds at matched magnitude and is the powered primary test, while the three-way ordering is declared undefined at matched magnitude rather than forced. The absence of common support is itself a recorded finding, not a failure to report.

\paragraph{The pullback ruler.} The construction has used length freely: the reach of the shadow segment $OP$, the value $\operatorname{mag}(h)$, and the shared targets $m_{\mathrm{low}}$ and $m_{\mathrm{high}}$. The ordinary Euclidean ruler is not the right one here, because the plane sits inside a model. What matters about $d_1$ and $d_2$ is not only how they look in activation space, but what the model's downstream computation does with them.

The model may stretch one direction more than another, shrink a direction, or shear the angle between them. Two directions that look orthogonal and equal in activation space need not remain orthogonal or equal after the downstream computation acts on them. The design therefore measures the plane's geometry as the model renders it downstream, then pulls that geometry back onto the plane. This gives the pullback metric: an inner product on the plane that records how the model sees its directions.

\paragraph{The Jacobian construction.} The pullback ruler is built from the model's local linear map. Let $J$ be the Jacobian of the downstream computation at the point being probed. Applying $J$ to the plane's two directions gives the restricted Jacobian:
\[
  JD = \begin{bmatrix} Jd_1 & Jd_2 \end{bmatrix}
\]
Its columns are the two plane directions as the model renders them downstream. The inner products of these two downstream images form the pullback metric:
\[
  M = (JD)^\top(JD)
\]
This is the same $M$ used in \Cref{sec:geometric-setup-instrument} for area and non-degeneracy. Here its job is to measure the shadow length used for magnitude matching; transport itself is still the restricted-Jacobian composition defined in \Cref{eq:transport}.

Writing $a$ for the projection coefficients,
\begin{equation}
\label{eq:in-plane-magnitude}
  a = M^{-1} JD^\top (Jh), \qquad
  \operatorname{mag}(h) = \lVert JD\,a \rVert_2 = \sqrt{a^\top M a}
\end{equation}
This is the same shadow length described above, now measured with the model's ruler rather than the ordinary one.

\paragraph{Pre-data revision.} The design was first frozen, then revised under a pre-data correction rule when a specification error was caught: the planning variance had been measured at an activation site that did not match the SAE feature dictionary used to build the planes. The dictionary is defined at the post-layer site, while the variance had been taken at the pre-layer site. The design was corrected to the matched site, the variance re-estimated on reserved points only, the required sample size re-derived, and the design re-frozen before the experiment sample was touched. Only the reserved pilot points were consumed by this correction.

\paragraph{Verdict rule and materiality.} A materiality threshold is applied to every hypothesis: an effect counts as material only if it reaches a $25\%$ multiplicative change in holonomy, equivalently $\log 1.25 = 0.2231$ on the log scale. Each hypothesis's confidence interval is then compared to this threshold under the four-way verdict rule of \Cref{sec:methodology}. The sample size was chosen in advance to detect a materiality-threshold-sized effect with $90\%$ power, using the corrected-site variance estimate. The analysis runs once; there is no peeking and no result-dependent stopping.

\paragraph{Primary test and covariates.} The primary test is a within-base-point paired comparison of active-feature versus mixed-feature holonomy on the log scale, with the two magnitude levels averaged within each plane type before comparing. The fitted primary model is ordinary least squares on the paired active-feature-minus-mixed-feature outcome, with centered covariate differences as adjustments; its intercept is the adjusted mean effect at the sample covariate centroid. The confirmatory interval is the two-sided 95\% OLS $t$-interval on that intercept, with an HC3 interval reported as the preregistered robustness check. The two nuisance quantities are measured per plane: an off-manifold proxy --- how poorly the model's own SAE feature dictionary reconstructs the loop's visited points --- and the angle between the two plane directions. The off-manifold proxy adjusted for here is reconstruction distance from the SAE feature dictionary; it is named precisely because the Discussion later distinguishes it from a different geometric notion of off-manifold displacement.

\paragraph{Off-manifold guard.} In addition to the reconstruction-distance adjustment, a second off-manifold proxy was preregistered as a guard. The preregistered first choice was a Mahalanobis distance from the distribution of observed activations, with a $k$-nearest-neighbour distance as the fallback if the covariance estimate was ill-conditioned. In this design the fallback was unavoidable: the reference covariance is estimated from $390$ activations in a $2304$-dimensional space, so after centering its rank is at most $389$, far below the ambient dimension, and it cannot support a stable Mahalanobis distance. The guard therefore used a $k$-nearest-neighbour distance for every plane --- the mean distance to the nearest reference activations, a measure of how isolated a point is from where activations normally lie. It was recorded as a balance diagnostic and was not part of the primary adjustment.

Two implementation details matter for the matched-center construction. First, the holonomy measurement does not reuse the Jacobian computed at the raw activation $h$ to measure a loop centered at $c$. The loop is built around $c$, and the restricted Jacobians are recomputed by Jacobian--vector products at the loop's own points. Second, the off-manifold proxies are evaluated on the displaced center and its swept loop points, not only on the original activation. Thus the locality and off-manifold checks attach to the region where holonomy is actually measured. They remain proxies, however: they can bound the concern, but they do not prove that the matched center is a natural activation or close the confound.

\paragraph{Secondary combined model.} In addition to the primary paired test, a combined mixed-effects model covers all three plane types and both magnitudes in one frame. On the log-holonomy scale it uses fixed effects for plane type, magnitude level, their interaction, reconstruction distance, and plane angle, plus a random intercept for base point. Its intervals are two-sided Wald intervals from the REML mixed-model fit; the preregistered fixed-effects fallback was not used because the random-intercept variance was not at the boundary. This model supplies the magnitude verdict and reports the ordering prediction descriptively. It also serves as a robustness check on the primary active-feature versus mixed-feature contrast; that corroborating estimate is reported with the other H-sem variants in the \hyperref[sec:results]{Results}.

\section{Results}
\label{sec:results}
% JOB: report the verdicts as facts; let the reversal speak; no interpretation yet.
Before any result was computed, the analysed data were verified to be byte-for-byte the same as the frozen measurement, with the full set of measurements present: 2,340 across 390 base points, and no invalid values. The results rest on the exact data that were locked before analysis.

\begin{table}[ht]
\centering
\small
\begin{tabular}{@{}p{0.17\linewidth}p{0.77\linewidth}@{}}
\toprule
\textbf{Prediction} & \textbf{Frozen result} \\
\midrule
\textbf{H-sem} & Falsified. The adjusted paired active-feature minus mixed-feature contrast was \(-0.29439\), with 95\% interval \([-0.43989,-0.14889]\), on the side opposite the prediction. \\
\textbf{H-mag} & Falsified. The high-minus-low matched-magnitude effect was \(-0.09367\), with 95\% interval \([-0.22675,0.03942]\). \\
\textbf{H-grad} & Undefined, no verdict. The random to mixed-feature to active-feature ordering was not estimated at matched magnitude because the three plane types had no three-way common support. \\
\bottomrule
\end{tabular}
\caption{\textbf{Frozen verdicts.} The semantic main effect (\textbf{H-sem}) was the powered primary test. The magnitude main effect (\textbf{H-mag}) came from the secondary combined model. The monotone-gradient hypothesis (\textbf{H-grad}) received no verdict because the required matched-magnitude comparison was unsupported.}
\label{tab:results-verdicts}
\end{table}

The central prediction, \textbf{H-sem}, was that active-feature planes would carry more holonomy than the matched mixed-feature controls. It failed, and failed in reverse: the active-feature planes carried less holonomy. This is a reversal, not an absence of effect (\Cref{tab:results-verdicts}).

\begin{figure}[ht]
\centering
\includegraphics[width=0.8\linewidth]{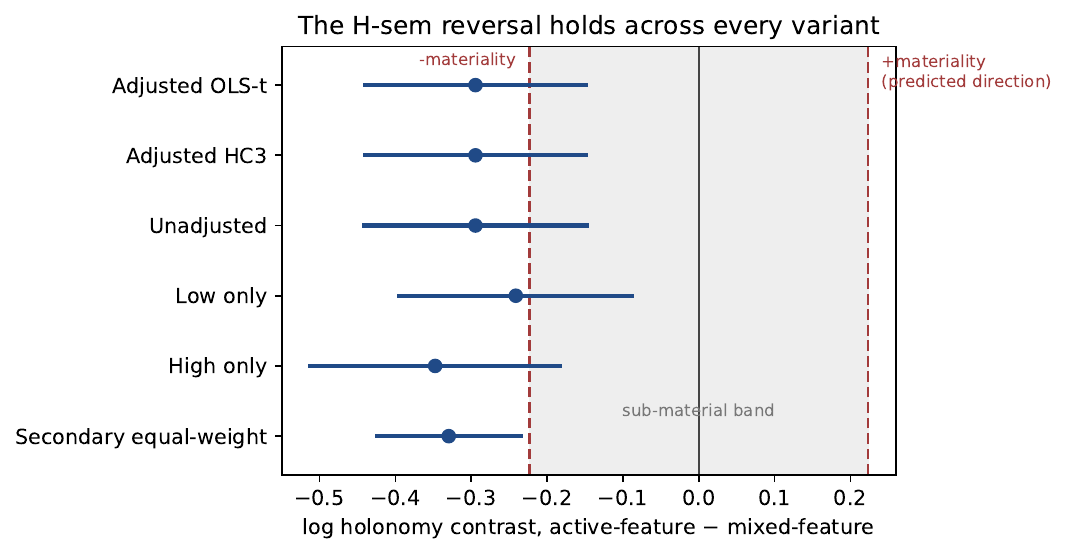}
\caption{\textbf{The H-sem reversal is robust across analyses.} The active-feature versus mixed-feature holonomy contrast (log scale) for six reported analyses: the confirmatory adjusted primary estimate, its unadjusted and HC3 robustness variants, the two within-magnitude descriptive contrasts, and the secondary equal-weight model. Every point estimate falls beyond the negative materiality threshold ($-0.2231$), and every confidence interval lies below zero. The right dashed line marks $+0.2231$, where the semantic-concentration prediction expected the effect; all variants sit on the opposite side.}
\label{fig:hsem-forest}
\end{figure}

The size of the reversal is about a quarter less holonomy on the active-feature planes. The adjusted paired estimate is \(-0.29439\) on the log scale, a 26\% multiplicative reduction. Its 95\% interval, \([-0.43989,-0.14889]\), lies entirely below zero and entirely below the pre-set materiality threshold of \(+0.2231\), so the result is on the opposite side of the threshold from the prediction. The point estimate also lies beyond the negative materiality threshold, \(-0.2231\). \Cref{fig:hsem-forest} shows the confirmatory estimate alongside robustness and descriptive variants.

The preregistered robustness checks agree with the confirmatory estimate. The unadjusted estimate is the same to the displayed precision, \(-0.29439\) with 95\% interval \([-0.44064,-0.14815]\). The robust HC3 interval, \([-0.44034,-0.14844]\), gives the same verdict. Adjusting for the two nuisance quantities leaves the estimate essentially unchanged.

One nuance must not be flattened: that the adjustment changed almost nothing is reported as a fact about the adjustment. It is not a claim that an underlying confound is ruled out. Whether the reversal reflects meaning or something else is left to the Discussion.

The reversal is present at both matched magnitude levels. At low magnitude the active-feature minus mixed-feature contrast is \(-0.24122\), with interval \([-0.39474,-0.08769]\). At high magnitude it is \(-0.34756\), with interval \([-0.51251,-0.18262]\). The secondary equal-weight model gives the same qualitative result, \(-0.32964\) with interval \([-0.42446,-0.23481]\). The two analyses are distinct computations and their numbers are kept labelled by model.

The secondary prediction, \textbf{H-mag}, that larger magnitude raises holonomy, also failed. The high-minus-low estimate is \(-0.09367\), with interval \([-0.22675,0.03942]\). Its interval includes zero and sits below the materiality threshold, so there is no material magnitude main effect in this design.

The third prediction, \textbf{H-grad}, a rising order across all three plane types, could not be tested because the three types did not share enough common ground to compare at matched magnitude. It is reported descriptively only and receives no verdict: undefined, not failed.

The result is not driven by any single unusual base point; no base point dominates. The secondary model fit cleanly without hitting a degenerate edge, and the active-feature and mixed-feature planes overlapped enough that the adjusted estimate is a genuine comparison rather than an extrapolation.

\begin{figure}[ht]
\centering
\includegraphics[width=0.9\linewidth]{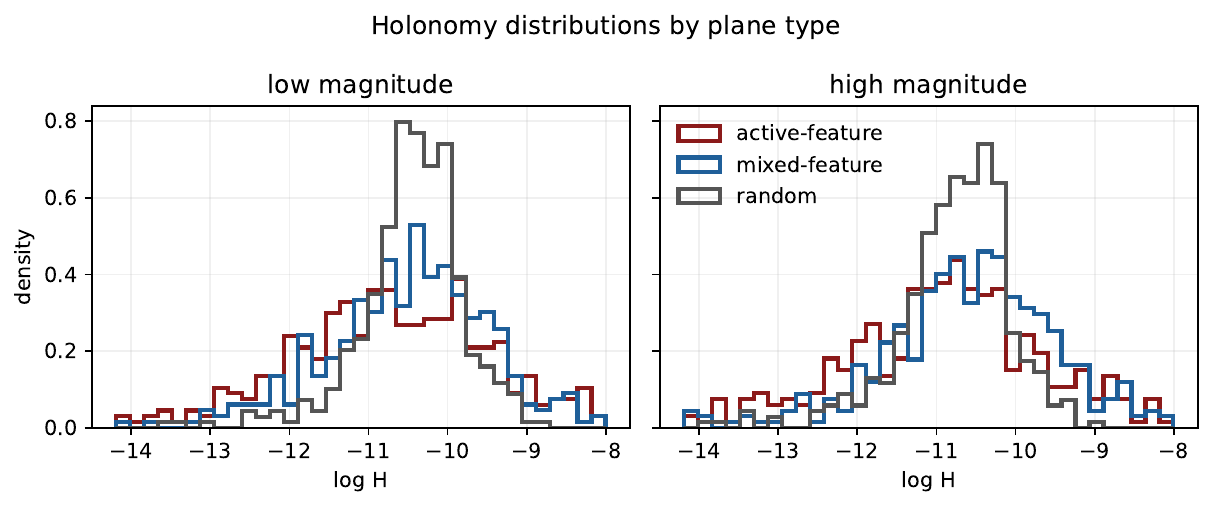}
\caption{\textbf{Holonomy distributions by plane type.} Per-plane-type distributions of log holonomy at low and high in-plane magnitude. The random planes are more sharply peaked than the active-feature and mixed-feature planes; the active-feature and mixed-feature distributions overlap substantially, even though the paired active-feature minus mixed-feature contrast is certified in the reverse direction.}
\label{fig:per-arm-distributions}
\end{figure}

The random planes are descriptive context only. In the secondary model, at low magnitude they carried more holonomy than the mixed-feature planes: \(0.19022\) on the log scale, with interval \([0.04040,0.34003]\). At high magnitude they carried about the same holonomy as the mixed-feature planes: \(-0.03561\), with interval \([-0.18543,0.11420]\). These descriptive contrasts are shown in \Cref{fig:per-arm-distributions}, but they carry no verdict for the same reason the three-way ordering is undefined.

\section{Discussion}
\label{sec:discussion}
% JOB: convert a sign flip into an opened question without over-claiming causality.
Read the result first as a sign reversal in a certified comparison: the study asked whether active-feature planes carry more holonomy than matched mixed-feature controls, and in this design they carried less.

A reversal is more informative than a null. A null would say only that the experiment failed to find concentration at the materiality scale; a reversal says the geometry does distinguish the two plane types, but in the direction opposite the motivating theory.

The reversal certifies an operational contrast, not a causal story. It does not show that meaning suppresses holonomy, or that the controls are ``more semantic'' than the SAE feature pairs. It shows that this instrument, at this site, under this matching rule, assigns lower holonomy to the active-feature planes than to the matched mixed-feature controls.

The central open question is therefore not whether the measured contrast is present, but what it tracks. Semantic organization remains one possible reading, but it is not the leading causal conclusion. The design also entangles semantic selection with activation strength, degree of feature engagement, dictionary geometry, proximity to the activation manifold, the angle between the plane directions, and channels the covariates did not capture.

Two alternatives to the semantic reading are built into how the planes are selected. The first is an activation-strength effect. An active-feature plane is chosen from the strongest active SAE feature pair at the base point. Directions the model drives strongly together may be directions it uses along smoother, lower-curvature parts of its geometry. The mixed-feature control, by contrast, introduces an inactive direction the model is not using there, which may cut transversely into higher-curvature regions. If so, part of the reversal would follow from selecting planes by activation strength, with no semantic content required.

The second is a degree-of-engagement effect. The mixed-feature control keeps one active SAE feature and replaces the other with an inactive one. The contrast is therefore not two meaningful directions versus none; it is two active SAE feature directions versus one active SAE feature direction paired with a dormant one. A mixed-feature plane is partly engaged, not meaningless, so describing the result as ``meaningful planes carry less holonomy'' would overstate the semantic contrast.

The substantial active-feature and mixed-feature distributional overlap in \Cref{fig:per-arm-distributions} is consistent with this caution: the mixed-feature control is not a baseline with semantic selection removed, but a partly engaged feature plane.

Because both alternatives are introduced by the construction itself, the result should be read operationally: this design cannot separate semantic content from activation-strength geometry or from the degree of feature engagement. This is the central reason the causal question stays open, and why the result is stated as a property of the active-feature versus mixed-feature construction rather than of meaning.

One simpler deflationary explanation is not supported here. Under the matched design there was no material magnitude main effect, so the hypothesis that holonomy merely tracks in-plane magnitude does not hold in this setting. This does not claim magnitude never matters; it says magnitude did not explain the reversal in this design.

The random planes should not be overread. Lacking common support at matched magnitude, they provide descriptive context and a lower geometric anchor, not a verdict on a three-way gradient.

The natural manifold objection remains important. The active-feature planes might be closer to the model's natural activation manifold, the mixed-feature planes farther away, and the instrument might assign less holonomy near the manifold than off it. If that is the true mechanism, the reversal would be geometric but not semantic.

The covariate adjustment is useful but not decisive. The two nuisance quantities, reconstruction distance from the SAE feature dictionary and the plane angle, were adjusted for, and the reversal barely moved. That is evidence the reversal is not explained by those two measured quantities; it is not proof that every manifold or geometry confound has been removed.

The two senses of ``off-manifold'' must be kept distinct. Reconstruction distance measures how well the SAE feature dictionary reconstructs the visited points; it is not the same object as proximity to the model's natural activation manifold, nor the curvature-relevant displacement governed by the downstream map's geometry. A null on the first is not a null on the second.

The two manifolds were also not measured equally well. Reconstruction distance, which tracks the SAE dictionary's reconstruction manifold, was available and adjusted for; the proxy intended for the model's natural activation manifold, a Mahalanobis distance, could not be computed because the $390$-point covariance in $2304$ dimensions has rank at most $389$ after centering, and degraded to a k-nearest-neighbour distance used only as a diagnostic. So the side of the distinction that matters most for the confound is the side this run measured least well, which is a further reason the causal question stays open.

As an exploratory check beyond the frozen design, that diagnostic guard distance was related directly to the holonomy gap. Forming, for each base point, the active-feature-minus-mixed-feature difference in the k-nearest-neighbour guard distance and regressing the paired holonomy gap on it, using centered ordinary least squares on the same scale as the primary contrast, yields a negligible association: an \(R^2\) of about \(0.009\), under one percent of the gap's variation, with a slope interval that includes zero. For reference, the reconstruction-distance difference explains about one tenth of one percent. Adjusting the primary contrast on the guard difference, in place of or in addition to reconstruction distance, leaves the contrast intercept unchanged at \(-0.29439\). At the plane level the holonomy/guard association is weak within the feature-based planes; only the random planes, which carry no verdict, show a modest descriptive association. This analysis is exploratory and reported as such; it does not enter the confirmatory verdict. The script and report are published in the experiment repository as \path{analysis/guard/exploratory_guard_regression.py} and \path{reports/guard_regression_exploratory.md}.

This strengthens the manifold-proximity rebuttal without closing the confound. Both off-manifold proxies measured here, reconstruction distance and the k-nearest-neighbour guard, fail to predict the holonomy gap, and adjusting on either leaves the contrast intact, so a reviewer attributing the reversal to manifold proximity must explain why neither proxy tracks it. But the guard is the degraded proxy, so the activation-manifold question is not fully settled; and neither proxy speaks to the activation-strength or degree-of-engagement alternatives. The confound stays open.

Two further post-freeze diagnostics characterize the instrument rather than changing the verdict. First, the matched centers are indeed constructed points, often many loop radii from the original activation. In the frozen manifest, the displacement \(\lVert c-h\rVert/\rho\) has median \(10.6\) and 95th percentile \(29.7\) for active-feature planes, and median \(13.9\) and 95th percentile \(52.3\) for mixed-feature planes. This makes the matched-center concern concrete: the loops are local around their constructed centers, but those centers are not guaranteed to be natural activations. A paired post-freeze regression then tested whether the active-feature-minus-mixed-feature displacement difference predicts the paired holonomy gap. The average-displacement model had \(R^2=0.00645\), and its slope interval included zero \([-0.00174,0.0163]\). Thus the displacement asymmetry is present and directionally unfavorable, but in this diagnostic it does not linearly explain the paired reversal. Second, because the transport operator is not assumed orthogonal, a full post-freeze shear diagnostic recomputed transport distortion on all 390 verdict-bearing base points for active-feature and mixed-feature planes at both matched magnitudes. The transport-residual-norm difference explained the largest share of paired-gap variation, with \(R^2=0.39530\) and an HC3 interval excluding zero \([7604.1,15819.1]\). The active-minus-mixed symmetric-residual difference explained an intermediate share, with \(R^2=0.14467\), but its HC3 interval included zero \([-10186.8,39097.9]\). Maximum condition number explained a smaller share, with \(R^2=0.08159\) and an HC3 interval excluding zero \([0.49052,1.34749]\). Transport shear and non-orthogonality are therefore live mechanisms or confounds for what the instrument measures, not nuisances bounded away by the present design.

The study's value is the reversal together with the discipline around it: an overturned hypothesis is useful only if the reader can see that the hypothesis, the materiality threshold, the design, and the analysis were fixed before the outcome was inspected.

The shear diagnostic admits two interpretations. Conservatively, shear and non-orthogonality threaten a clean reading of the antisymmetric component of transport as pure rotation: the active/mixed contrast may partly reflect non-rotational distortion in the transport operator. More constructively, the same quantities may be informative geometric features of the downstream map. Active SAE feature planes may lie in regions where local transport is more coherent, while mixed-feature planes induce more stretch, shear, or frame non-orthogonality. The present design cannot distinguish those interpretations, because shear was diagnosed after the confirmatory run rather than controlled by design.

The next experiment should therefore be a separation protocol, not an attempt to rescue the prediction by relabelling these numbers. First, match the control's activation strength to the active-feature plane, rather than drawing a dormant partner, so that strength-ranked selection cannot masquerade as meaning. Second, hold fixed the number of active SAE feature directions, for instance by comparing the selected active-feature pair with another two-active pair not chosen as the strongest pair. Third, match curvature-relevant displacement, so that proximity to the activation manifold cannot masquerade as the effect. Fourth, match or stratify by transport shear and non-orthogonality, report rotation-only variants of the holonomy readout, and test whether shear metrics predict independent measures of semantic coherence. Each prong isolates one factor the present design entangles.

\section{Limitations and Threats to Validity}
% JOB: enumerate the scope boundaries and threats to interpretation without minimizing them.
Scope. The claim is bounded to a single setting: one model family and size, one layer site, one SAE dictionary, one readout, one transport rule, one loop-radius rule, and one pullback-metric response. It is not a universal statement about transformer geometry.

Scope, corpus. The corpus is English natural prose; the result need not transfer to dialogue, code, multilingual text, instruction data, or other activation regimes.

Construct validity, behaviour. The study measures a local geometric quantity, not behaviour. It does not show that changing holonomy changes the model's outputs, that holonomy has causal force, or that the reversal affects task performance.

Construct validity, the semantic label. The semantic label attaches to how the planes are built, not to a behavioural assay. Active-feature planes are built from two active SAE feature directions, while mixed-feature planes pair one active SAE feature with one inactive SAE feature; this is a principled semantic proxy, but a proxy.

Confound, manifold control. The manifold control is incomplete by design. The adjusted proxy, reconstruction distance from the SAE feature dictionary, is not the same object as proximity to the model's natural activation manifold or the curvature-relevant displacement governed by the downstream map. A point can reconstruct well yet be geometrically unusual, so this confound is bounded but not closed.

Matched-center displacement. Magnitude matching constructs a new loop center $c$ by rescaling the activation's in-plane component while preserving its out-of-plane component. For the verdict-bearing active-feature and mixed-feature comparison, the matched targets sit inside native magnitude support, the holonomy Jacobians are recomputed around $c$, and the off-manifold proxies are evaluated on $c$ and its swept loop points. Those facts bound the concern, but they do not eliminate it. The center $c$ is still a constructed point, the displacement from the raw activation can be many loop radii, and the preregistration did not separately analyse displacement relative to $\rho$ or provide a binary test that $c$ is a natural activation.

The post-freeze displacement diagnostic quantifies this limitation. In units of loop radius, the active-feature centers had median displacement \(10.6\), 95th percentile \(29.7\), and maximum \(87.8\); mixed-feature centers had median \(13.9\), 95th percentile \(52.3\), and maximum \(145\). Random-plane centers were farther still, but random planes carry no matched-magnitude verdict. A paired displacement-gap regression further bounded the concern: active-minus-mixed displacement difference explained less than one percent of paired-gap variation (\(R^2=0.00645\)), with a slope interval crossing zero. This weakens a simple linear displacement explanation, but it does not close nonlinear, interaction, or natural-manifold explanations.

One preregistered guard could not run as specified. The second off-manifold proxy was planned as a Mahalanobis distance, but the activation covariance has rank at most $389$ after centering in this $390$-point, $2304$-dimensional regime, so the guard degraded to a k-nearest-neighbour distance. The measure intended to capture distance from the natural activation distribution is therefore the weaker of the two off-manifold proxies in this run.

Design coverage. The lack of common support limits one part of the design. Because the three plane types did not share enough matched magnitude range, the three-way ordering is undefined at matched magnitude, and any pattern involving random planes is descriptive context, not a verdict.

Instrument. The instrument is validated but remains an instrument. Synthetic and oracle checks, and a post-freeze same-readout radius sweep, support the transport logic and area-law behaviour, but the measurement is still tied to the chosen restricted-Jacobian transport rule, layer-12-to-layer-13 readout, radius, loop discretization, and the bounded backend stated in Methodology: MPS execution plus deterministic CPU float64 small-matrix arithmetic. The full post-freeze shear diagnostic reports that transport distortion co-varies with the paired gap --- most strongly for the aggregate residual norm (robust), weakly but also robustly for non-orthogonality, and, for the symmetric-residual shear component, materially under conventional intervals but not under robust ones. This does not alter the frozen operational reversal, but it makes transport distortion a live mechanism or confound for its interpretation.

Integrity and disclosure. The single pre-data design correction is a strength but also a disclosure obligation. The design was frozen, a site mismatch was caught and corrected before any experiment data were observed using only reserved points, and the design was re-frozen; the paper's integrity rests on that record, which is committed with full version history and publicly auditable in the released experiment repository. The correction is detailed in the Experimental Design chapter.

Inferential scope. The confidence intervals and verdicts answer the preregistered materiality question, not every question one might ask of the data. Smaller effects, other contrasts, and alternative geometries are future studies, not hidden conclusions of this one.

\section{Conclusion}
% JOB: restate the ladder: measurable holonomy, falsified semantic concentration, narrow certified contrast, open cause, next experiment.
The answer to the title question is no in this preregistered Gemma setup: active SAE feature planes carried less measured holonomy than matched mixed-feature controls, not more.

The motivating prediction was falsified in reverse: active-feature planes did not carry more holonomy than matched mixed-feature controls; the point estimate lay in the opposite, materially negative direction, about a 26\% reduction.

The reversal is credible because the hypothesis, materiality threshold, design, and analysis were fixed before the outcome was inspected, the analysed data were verified before estimation, and the verdict rule was applied as written.

The certified claim is narrow and not deflationary in every direction. It does not say geometry is irrelevant to representations; it says this specific semantic-concentration prediction failed, in reverse, under this specific frozen measurement.

The magnitude-only explanation was not supported in this design, and the three-way ordering remains undefined rather than failed.

What the reversal tracks remains open: activation-strength geometry, degree of feature engagement, dictionary geometry, matched-center displacement, proximity to the activation manifold, transport shear, semantic organization, or another channel of the model-induced geometry.

The next experiment should therefore be designed not to ask whether the original prediction can be saved, but to separate those explanations cleanly.

\section*{Use of AI tools}

This project made substantial use of large language model assistants --- Anthropic's Claude, OpenAI's ChatGPT (including its Codex coding scaffolding), and Google's Gemini --- which the author directed as specialist collaborators across system and experimental design, code implementation and review, and manuscript preparation, at times collaboratively and at times to cross-check one another. In this workflow the author acted as orchestrator and retained every decision; concretely, the tools were used for design discussion, developing and refactoring analysis and figure-generation code, adversarial code and result review, copy-editing and restructuring exposition, and checking the internal consistency of reported figures. The scientific commitments that make this a confirmatory study --- the hypotheses, the materiality threshold, the design, and the analysis --- were set by the author and preregistered and frozen before the analysed measurements were inspected, and all reported numerical results were produced by the author from the frozen experiment repository. The author verified every quantitative claim and every reference against that record and takes full responsibility for the entire contents of this paper, including any content produced with the assistance of these tools. No AI tool is an author of this work.

\appendix

\section{Frozen Record and Corpus Construction}
\label{app:frozen-record-corpus}

\paragraph{Public experiment record.} The source of truth for the confirmatory experiment is the public experiment repository, \path{gemma-holonomy-doe}. The archived release cited by this paper is \href{https://doi.org/10.5281/zenodo.20960027}{doi:10.5281/zenodo.20960027}; it is the frozen confirmatory record. The manuscript repository contains the paper text and paper-facing figure pipeline; it is not the frozen experimental record. The experiment repository contains the preregistration history, the run environment, the corpus draw, the frozen manifest, the frozen measurement table, the integrity receipt, and the analysis reports used by this paper. Large artifacts are tracked through Git LFS where needed: the Zenodo source archive records the LFS pointer for \path{run/manifest_n390.json}, while GitHub LFS serves the full manifest payload, whose full-file SHA-256 is recorded in the integrity reports.

The primary frozen artifacts are:
\begin{itemize}
\item \path{PREREGISTRATION.md} and \path{PREREGISTRATION_v2.md};
\item \path{ENVIRONMENT.md};
\item \path{run/corpus_draw_n390.json} and \path{run/manifest_n390.json};
\item \path{run/holonomy_results_n390.parquet} and \path{run/holonomy_results_n390.json};
\item \path{reports/integrity_receipt.json};
\item \path{reports/hsem_results.json} and \path{reports/hsem_results.md}.
\end{itemize}
The small result files copied into the paper repository are derived from those artifacts; they are conveniences for manuscript figures and tables, not a second experimental record.

The exploratory guard-regression check discussed in the Discussion is published in the public experiment repository as \path{analysis/guard/exploratory_guard_regression.py}, \path{reports/guard_regression_exploratory.json}, and \path{reports/guard_regression_exploratory.md}. It is a post-freeze exploratory analysis artifact, not part of the confirmatory frozen verdict record.

The post-freeze instrument and diagnostic checks discussed above are published in the public experiment repository:
\begin{itemize}
\item \path{analysis/validation/area_law_layer13.py};
\item \path{analysis/diagnostics/center_displacement.py};
\item \path{analysis/diagnostics/displacement_gap_regression.py};
\item \path{analysis/diagnostics/full_transport_shear.py};
\item \path{analysis/diagnostics/shear_gap_regression.py};
\item \path{analysis/diagnostics/transport_shear.py};
\item \path{reports/area_law_layer13_validation.*};
\item \path{reports/center_displacement_diagnostic.*};
\item \path{reports/displacement_gap_regression.*};
\item \path{reports/full_transport_shear_diagnostic.*};
\item \path{reports/shear_gap_regression.*};
\item \path{reports/transport_shear_diagnostic.*}.
\end{itemize}
They characterize the instrument and residual threats to validity; they do not enter or modify the preregistered decision rule.

\paragraph{Corpus and prompts.} The experiment does not use instruction prompts, chat templates, system prompts, or sampled model completions. The model inputs are raw natural-language passage prefixes used to elicit residual-stream activations. In this paper, the word ``prompt'' therefore means a fixed corpus passage prefix, not an instruction to the model.

The corpus is WikiText-103 raw with these identifiers:
\begin{itemize}
\item dataset: \path{Salesforce/wikitext};
\item configuration: \path{wikitext-103-raw-v1};
\item revision: \path{b08601e04326c79dfdd32d625aee71d232d685c3}.
\end{itemize}
WikiText raw is distributed as line-level rows, so the corpus draw reconstructs articles by joining lines between top-level article headers. Each reconstructed article is one candidate passage. Candidate passages are drawn with seed \(42\), tokenized with the \path{google/gemma-2-2b} tokenizer, and truncated to the first \(64\) tokens. Passages shorter than \(64\) tokens are dropped rather than padded.

\paragraph{Base points.} A base point is the residual-stream activation at the final token of one retained \(64\)-token passage prefix. Activations are captured at \path{resid_post} after block~12, the output of \path{model.model.layers[12]}. The corpus draw retains a \(700\)-passage pool after the preregistered exclusions and burned pilot draw orders; the experiment sample is the first \(390\) surviving base points in seed order. Each base point is evaluated under all planned plane-type and magnitude combinations, yielding \(390\times3\times2=2340\) measurements.

\paragraph{Model, SAE, and site.} The model is Gemma~2 2B, \path{google/gemma-2-2b}. The SAE dictionary is the Gemma Scope residual-stream SAE \path{google/gemma-scope-2b-pt-res}, path \path{layer_12/width_16k/average_l0_82/params.npz}. Plane directions are normalized decoder rows from that SAE dictionary at the same layer site. The \path{average_l0_82} suffix denotes an average L0 of about \(82\) active SAE features in the dictionary's operating regime.

\paragraph{Reproducibility boundary.} The confirmatory run is specified for Apple Silicon MPS, with fixed seeds and recorded artifacts. Model execution and Jacobian--vector products run on MPS; the small pullback matrices used for determinant, area, and projection calculations are evaluated in CPU float64. The reproducibility claim is therefore the one stated in the experiment repository: bitwise within the recorded MPS environment, with the frozen artifacts and integrity receipt as the audit trail.

\section{Glossary}

\glossentry{activation-manifold}{Activation manifold}
The activation manifold is the lower-dimensional region of \hyperref[gloss:activation-space]{activation space} that the model's natural activations actually occupy. It is distinct from reconstruction by the \hyperref[gloss:sae-feature-dictionary]{SAE feature dictionary}: a point can reconstruct well yet still be unusual relative to the model's natural activation manifold.

\glossentry{activation-space}{Activation space, residual stream}
Activation space is the high-dimensional space of the model's internal residual-stream activations, where each point is one activation vector. This is the space whose geometry the paper measures.

\glossentry{base-point}{Base point}
A base point is one raw activation vector in the model's \hyperref[gloss:activation-space]{residual-stream activation space}. The experiment uses it to choose plane directions, compute in-plane magnitude, and construct the matched loop centers. In this study each base point comes from the same fixed corpus-sampling rule, and every base point is measured under all planned plane-type and magnitude combinations.

\glossentry{active-feature}{Active feature}
An active feature is an SAE feature whose \hyperref[gloss:sae-code]{SAE code} is positive at the base point, meaning the sparse autoencoder assigns that feature a nonzero positive contribution there.

\glossentry{common-support}{Common support and two-plane-type fallback}
Common support means a range of in-plane magnitudes where the plane types being compared all have observed measurements, so matching does not require extrapolating beyond one plane type's native range. When the three plane types did not share enough common support, the preregistered two-plane-type fallback kept the active-feature versus mixed-feature comparison at matched magnitude and declared the three-way ordering undefined at matched magnitude.

\glossentry{curvature}{Curvature}
Curvature is, informally, how much the \hyperref[gloss:pullback-metric]{pullback geometry} bends locally. \hyperref[gloss:holonomy]{Holonomy} is the loop-integral signature of curvature: a nonzero twist around a small loop indicates curvature enclosed by that loop.

\glossentry{downstream-map}{Downstream map}
The downstream map is the part of the model used to say how a small change in activation space affects what comes later. In this experiment it is the final-token map from a patched layer-12 residual-stream activation to the layer-13 residual-stream output. Its local Jacobian induces the \hyperref[gloss:pullback-metric]{pullback metric}, which is the geometry used for distances, areas, and transport in this paper.

\glossentry{frame-parallel-transport}{Frame and parallel transport}
A frame is a small set of reference directions carried along a path so that the experiment can ask whether it returns rotated. Parallel transport is the rule for carrying that frame around the closed loop; the resulting rotation is the \hyperref[gloss:holonomy]{holonomy}.

\glossentry{holonomy}{Holonomy}
Holonomy is the twist a frame picks up when it is carried around a closed loop. In this experiment, holonomy is measured by tracing a small loop around the matched center associated with a base point at the layer-12-to-layer-13 readout and extracting the rotation of the transported frame when it returns; nonzero holonomy is the observed signature of local \hyperref[gloss:curvature]{curvature}.

\glossentry{holonomy-magnitude}{Holonomy magnitude}
Holonomy magnitude is the orientation-independent response used for the primary comparisons: the absolute rotation angle divided by the loop's enclosed area. The signed version is retained as a diagnostic, but the reported response uses magnitude so that reversing loop direction does not change the measured amount of twist; area-normalizing makes it a local \hyperref[gloss:curvature]{curvature} measure.

\glossentry{in-plane-magnitude}{In-plane magnitude}
In-plane magnitude is the size of a base-point activation's component within the two-dimensional plane being probed, measured under the \hyperref[gloss:pullback-metric]{pullback metric}. The experiment uses two matched in-plane magnitude settings so that plane types are compared at the same size scale.

\glossentry{inactive-feature}{Inactive feature}
An inactive feature is an SAE feature whose \hyperref[gloss:sae-code]{SAE code} is zero at the base point, meaning the sparse autoencoder does not use that feature in its sparse representation there.

\glossentry{materiality-threshold}{Materiality threshold}
The materiality threshold is the \hyperref[gloss:preregistration]{preregistered} bar an effect must clear to count as meaningfully large. Here the bar is a 25\% multiplicative change in holonomy, equal to \(0.2231\) on the log scale.

\glossentry{off-manifold-proxy}{Off-manifold proxies}
The paper uses two off-manifold proxies, which must not be collapsed. The primary adjusted proxy is reconstruction distance from the sparse-autoencoder feature dictionary: it measures how poorly the dictionary reconstructs the loop points being visited. The preregistered guard was intended to measure distance from the model's natural \hyperref[gloss:activation-manifold]{activation manifold}; in this run it was computed as a \hyperref[gloss:knn-guard]{k-nearest-neighbour guard distance}, used diagnostically rather than in the primary adjustment.

\glossentry{knn-guard}{k-nearest-neighbour guard distance}
The k-nearest-neighbour guard distance is the diagnostic off-manifold score computed for distance from the observed activation distribution. It is the mean distance from a point to its nearest reference activations; it replaced the planned Mahalanobis guard because the activation covariance has rank at most \(389\) after centering in the \(390\)-point, \(2304\)-dimensional setting. It is a density or isolation score, not a clustering label or a classifier.

\glossentry{phi}{Phi, plane angle}
Phi is the ordinary Euclidean angle between the two raw directions that span a plane. It is measured because the learned angle between SAE feature directions may itself carry structure and could otherwise be confused with a semantic effect.

\glossentry{plane-types}{Plane types: active-feature, mixed-feature, and random}
An active-feature plane is spanned by two SAE features that are both \hyperref[gloss:active-feature]{active} at the base point, chosen as the strongest usable such pair under the non-degeneracy floor. A mixed-feature plane, the main control, pairs one active feature with one \hyperref[gloss:inactive-feature]{inactive feature}, the inactive partner drawn at random; it is built independently of the active-feature plane. A random plane is spanned by two random directions in activation space rather than SAE features, serving as a geometric floor.

\glossentry{pullback-gram-detm}{Pullback Gram and non-degeneracy floor}
The pullback Gram matrix \(M\) is the two-by-two matrix of inner products between the plane directions after they are viewed through the downstream map's Jacobian. Its determinant, \(\det(M)\), measures whether the plane remains non-degenerate under the pullback geometry; the non-degeneracy floor rejects planes with \(\det(M)\leq0.413\) before holonomy is measured.

\glossentry{pullback-metric}{Pullback metric}
The pullback metric is the geometry on \hyperref[gloss:activation-space]{activation space} induced by the downstream map: two activation-space directions are close or far according to how their downstream effects compare. This paper uses that metric for loop area, in-plane magnitude, and transport rather than using ordinary Euclidean geometry.

\glossentry{preregistration}{Preregistration}
Preregistration means fixing the hypotheses, decision rule, materiality threshold, design, and analysis in writing, then freezing them before the analysed measurements are inspected. It is the basis for treating a result as confirmatory rather than chosen after the fact.

\glossentry{sae-code}{SAE code}
The SAE code is the vector of coefficients the sparse autoencoder assigns to an activation, one coefficient per dictionary feature. Most coefficients are zero; an SAE feature is active where its code is positive and inactive where its code is zero.

\glossentry{sae-feature-dictionary}{Sparse-autoencoder feature dictionary}
A sparse-autoencoder feature dictionary is a learned set of SAE feature directions used to represent model activations as sparse combinations of SAE features. In this study the active-feature and mixed-feature planes are built from directions in the Gemma Scope SAE dictionary at the measured layer site. Sevetlidis and Pavlidis use ``feature'' for raw activations or PCA directions rather than SAE features; this paper's SAE features are learned dictionary directions from a sparse autoencoder.

\glossentry{semantic-concentration-prediction}{Semantic-concentration prediction}
The prediction this paper tests: that holonomy concentrates on meaningful directions --- that \hyperref[gloss:plane-types]{active-feature planes} carry materially more holonomy than matched mixed-feature controls. The preregistered test reversed it: active-feature planes carried materially less.

\glossentry{verdicts}{Verdicts: corroborated, falsified, inconclusive, and undefined at matched magnitude}
A prediction is corroborated if its confidence interval lies entirely above the \hyperref[gloss:materiality-threshold]{materiality threshold}, falsified if it lies entirely below that threshold, and inconclusive if it straddles the threshold. Undefined at matched magnitude means the relevant comparison could not be given a verdict because the required matched-magnitude comparison was not supported by the data. These verdict rules were fixed by \hyperref[gloss:preregistration]{preregistration}.


\begin{thebibliography}{9}

\bibitem{bricken2023towards}
T. Bricken et al.
\newblock Towards monosemanticity: Decomposing language models with dictionary learning.
\newblock \emph{Transformer Circuits Thread}, 2023.
\newblock \url{https://transformer-circuits.pub/2023/monosemantic-features/}.

\bibitem{elhage2022toy}
N. Elhage et al.
\newblock Toy models of superposition.
\newblock \emph{Transformer Circuits Thread}, 2022.
\newblock \url{https://transformer-circuits.pub/2022/toy_model/}.

\bibitem{gemma2024gemma2}
Gemma Team.
\newblock Gemma 2: Improving open language models at a practical size.
\newblock arXiv:2408.00118, 2024.
\newblock \url{https://arxiv.org/abs/2408.00118}.

\bibitem{lieberum2024gemma}
T. Lieberum, S. Rajamanoharan, A. Conmy, L. Smith, N. Sonnerat, V. Varma, J. Kramar, A. Dragan, R. Shah, and N. Nanda.
\newblock Gemma Scope: Open sparse autoencoders everywhere all at once on Gemma 2.
\newblock In \emph{Proceedings of the 7th BlackboxNLP Workshop: Analyzing and Interpreting Neural Networks for NLP}, pages 278--300. Association for Computational Linguistics, 2024.
\newblock \url{https://aclanthology.org/2024.blackboxnlp-1.19/}.

\bibitem{merity2017pointer}
S. Merity, C. Xiong, J. Bradbury, and R. Socher.
\newblock Pointer sentinel mixture models.
\newblock In \emph{International Conference on Learning Representations}, 2017.
\newblock \url{https://openreview.net/forum?id=Byj72udxe}.

\bibitem{sevetlidis2026gauge}
V. Sevetlidis and G. Pavlidis.
\newblock Gauge-invariant representation holonomy.
\newblock In \emph{International Conference on Learning Representations}, 2026.
\newblock \url{https://openreview.net/forum?id=czJqKToDGq}.

\bibitem{vaswani2017attention}
A. Vaswani, N. Shazeer, N. Parmar, J. Uszkoreit, L. Jones, A. N. Gomez, L. Kaiser, and I. Polosukhin.
\newblock Attention is all you need.
\newblock In \emph{Advances in Neural Information Processing Systems}, 2017.
\newblock \url{https://arxiv.org/abs/1706.03762}.

\end{thebibliography}
\end{document}